%% file: main.tex
\documentclass[11pt]{article}

\usepackage[final]{acl}
\usepackage{graphicx}
\usepackage{caption}
\usepackage{adjustbox} % keeps image inside page 
\usepackage{booktabs}
\usepackage{siunitx} % 用于小数点对齐，非常关键
\usepackage{multirow} % 用于多行单元格，可选
\usepackage{float}      % 提供 [H] 选项，强制将浮动体放在“这里”
\usepackage{placeins}   % 提供 \FloatBarrier 命令，防止浮动体乱跑
\usepackage{times}
\usepackage{latexsym}
\usepackage{booktabs}
\usepackage{tabularx}
\usepackage{array}
\usepackage{enumitem}
\usepackage{amsmath}    % 基础数学公式结构（向量、上下标、集合）
\usepackage{amssymb}    % 补充数学符号（黑板粗体$\mathbb{R}$、花体$\mathcal{L}$）
\usepackage[T1]{fontenc}
\usepackage[utf8]{inputenc}

\usepackage{microtype}

\usepackage{inconsolata}

\usepackage{graphicx}
\usepackage{amsthm}

\newtheorem{definition}{Definition}
\newtheorem{hypothesis}{Hypothesis}

\usepackage{algorithm}
\usepackage{algorithmic}  % 或改为 \usepackage{algpseudocode}
\title{Hierarchical Alignment: Block-Level LoRA-DPO for Preference Tuning in Large Language Models}
\hypersetup{
  pdftitle={Hierarchical Alignment: Block-Level LoRA-DPO for Preference Tuning in Large Language Models},
  pdfauthor={Yishen Chen, Qi Dong, and Yukun Zhang}
}

\author{
  Yishen Chen\textsuperscript{*} \\
  \normalfont The Chinese University of \\
  Hong Kong, Shenzhen \\
  \texttt{yishenchen@link.cuhk.edu.cn}
  \And
  Qi Dong\textsuperscript{*} \\
  \normalfont Fudan University \\
  \texttt{19210980065@fudan.edu.cn}
  \And
  Yukun Zhang\textsuperscript{*} \\
  \normalfont The Chinese University of \\
  Hong Kong \\
  \texttt{215010026@link.cuhk.edu.cn}
}

\begin{document}
\maketitle
\begingroup
\renewcommand{\thefootnote}{\fnsymbol{footnote}}
\footnotetext[1]{Equal contribution.}
\endgroup
\begin{abstract}
Preference tuning typically distributes adapter updates across every Transformer layer. We study how the placement of a fixed LoRA budget affects Direct Preference Optimization (DPO). \emph{Hierarchical Alignment} partitions decoder depth into \emph{Local}, \emph{Intermediate}, and \emph{Global} regions and restricts LoRA updates to a chosen contiguous block. Across Qwen and Llama models, three-seed sliding-window sweeps reveal broad low-loss basins around intermediate depth. Under exact trainable-parameter matching, Middle achieves lower held-out DPO loss than an all-depth allocation on Qwen2.5-7B and Llama-3.1-8B. Generation-level evidence is more variable. LLM judges and a blinded human study leave Selected versus All unresolved on Qwen; on Llama, two pairwise judges favor All while the human interval includes zero. Trained-adapter interventions show that learned depth components affect the optimization objective; removing the Llama Selected window also lowers generation quality. Across Qwen2.5 at 3B, 7B, and 14B, Middle versus All remains unresolved under Qwen3-Max; at 3B and 14B, both improve over Base. In our implementation, Middle also uses less peak memory and takes shorter steps than a parameter-matched all-depth allocation. Together, the results support intermediate depth as a practical starting point for fixed-budget preference tuning, though the best window and its behavioral value still need per-setting validation.
\end{abstract}

\section{Introduction}

Direct Preference Optimization (DPO) provides a stable objective for aligning Large Language Models (LLMs) with preference data \citep{rafailov2024directpreferenceoptimizationlanguage}. Parameter-efficient implementations commonly place LoRA adapters at every Transformer layer. This convention leaves the depth allocation of the trainable budget largely unexamined. Probing studies find substantial variation in representations across depth \citep{van_Aken_2019,nadipalli2025layerwiseevolutionrepresentationsfinetuned}, motivating a controlled test of whether preference optimization is also sensitive to adapter placement.

We ask: \emph{where should a fixed LoRA budget be placed for preference tuning?} \emph{Hierarchical Alignment} divides decoder depth into three positional regions---\textit{Local} (shallow), \textit{Intermediate} (middle), and \textit{Global} (deep)---and applies LoRA-DPO \citep{hu2021loralowrankadaptationlarge} to a selected block. These labels specify depth ranges and remain agnostic about exclusive linguistic or reasoning functions. We evaluate the depth-thirds partition with sliding windows (three training seeds each), exact-capacity all-depth controls, blinded human evaluation, trained-adapter interventions, a 3B/7B/14B scale study, external LLM judges, and efficiency profiling. This design separates \emph{where} capacity is placed from \emph{how much} is available.

\paragraph{Findings.} Three-seed window sweeps produce broad low-loss basins around intermediate depth for Qwen2.5-7B and Llama-3.1-8B, with model-specific minima and frozen representative windows. Exact-capacity all-depth adapters have higher held-out DPO loss than the Intermediate block on both models. The objective-level difference between the selected window and the larger All configuration is unresolved. Behavioral results depend on the model. On Qwen2.5-7B, all four evaluators---three LLM judges and a blinded human panel---leave Selected versus All unresolved. On Llama, the two pairwise judges favor All; the human point estimate agrees but its interval includes zero. At 3B and 14B, both Middle and All improve over Base under Qwen3-Max and remain unresolved against each other. Adapter removals affect the objective; removing Llama's Selected window also lowers generation quality.

\pagebreak
\paragraph{Contributions.}
\begin{itemize}[topsep=2pt,itemsep=1pt,leftmargin=*]
\item We formulate preference tuning as a fixed-budget layer-allocation problem and implement contiguous-block LoRA-DPO with a coarse-to-fine depth search (\S\ref{sec:methodology}).
\item We separate objective-level and behavioral evidence using multi-seed windows, an exact-capacity all-depth baseline, external and human judges, adapter interventions, and a 3B/7B/14B scale study (\S\ref{sec:Experiment}).
\item We report measured parameter, adapter, optimizer-state, peak-memory, and step-time costs, together with layer-wise gradients, updates, and representation drift (\S\ref{sec:mechanism}).
\end{itemize}

The evidence supports intermediate depth as a practical candidate region for fixed-budget LoRA-DPO, most clearly in the optimization objective and the matched-capacity comparison. Whether this objective advantage translates to better generations must be evaluated separately for each model scale, evaluator, and baseline allocation.

Three comparisons organize the study. Middle versus Matched isolates 
\emph{placement} while holding the trainable parameter count exactly fixed. Selected versus All compares the window returned by the frozen search rule with the conventional, higher-capacity all-layer recipe. Middle versus Base at 3B, 7B, and 14B measures whether the restricted allocation produces a useful preference-tuning effect across scale. Separating these comparisons keeps placement at fixed capacity, the search outcome, and behavioral utility across scale from being treated as one claim.

The experimental workflow follows the same separation. Teacher-forced DPO loss on a frozen development split selects a window and quantifies objective effects across three training seeds. Generation from the seed-42 checkpoints is then evaluated on disjoint prompts by multidimensional, pairwise, programmatic, and human evaluators. Trained-adapter interventions test whether the learned depth allocation contributes to the observed effects. The resulting evidence supports a practical search procedure and shows where objective and behavioral rankings diverge.

\section{Related Work}
\label{sec:related}

Our work sits at the intersection of preference alignment for LLMs and layer-specific, parameter-efficient tuning.

\subsection{Preference Alignment for LLMs}

Modern LLM alignment has moved from Supervised Fine-Tuning to RLHF \citep{ouyang2022traininglanguagemodelsfollow} and to simplified offline alternatives such as DPO \citep{rafailov2024directpreferenceoptimizationlanguage} and SimPO \citep{meng2024simpo}. Recent work also studies how preference-data consistency affects multilingual DPO \citep{zhang2025cmalignconsistencybasedmultilingualalignment}. These methods typically apply the alignment loss to all trainable layers without examining how the update budget is distributed across depth. We treat this distribution as an explicit design variable.

\subsection{Layer-Specific and Parameter-Efficient Tuning}

A separate line of work argues that the importance of individual layers or modules is highly non-uniform, and that selective updates can be both efficient and effective.

\emph{Coarse-to-fine module selection.} CoFiTune \citep{zhang2024balancing} first performs coarse-grained tree search over modules and then applies fine-grained soft-masking, demonstrating that ``which modules to update'' is a non-trivial design choice in its own right.

\emph{Sparse layer updates.} LISA \citep{pan2024lisa} randomly freezes most layers during training and shows that sparse layer updates can match or exceed full fine-tuning under matched memory budgets. IST \citep{yao2024layer} learns explicit layer-importance scores and identifies random sparse updates as a strong baseline for informed selection.

\emph{Single-layer or single-head steering.} CogSteer \citep{wang2025cogsteer} steers behavior with interpretability-driven single-layer interventions and reports that middle layers are often more effective intervention points than the final layer. \citet{skean2025layer} argue that for many downstream tasks, intermediate-layer representations are more informative than the final layer.

\emph{Modular adapters at the value/persona level.} \textsc{Aligner} \citep{ji2024alignerefficientalignmentlearning} and Modular Pluralism \citep{feng2024modularpluralismpluralisticalignment} compose small external experts for fine-grained behavioral control without modifying the base model. LoRA \citep{hu2021loralowrankadaptationlarge} provides the parameter-efficient mechanism we adopt; its formulation is agnostic to which layers receive updates.

\emph{Probing evidence about Transformer depth.} \citet{van_Aken_2019} and \citet{nadipalli2025layerwiseevolutionrepresentationsfinetuned} document that early layers capture syntax while middle and late layers encode semantics and task-specific logic. Related depth-localized effects have also been observed for factual recall in Mamba \citep{sharma2024locatingeditingfactualassociations}. These results motivate our depth partition. Unlike probing studies, our experiments directly test whether preference learning is sensitive to where updates are placed across depth.

\subsection{Positioning}

Our contribution is a controlled study of \emph{preference alignment} under contiguous block-level updates and matched LoRA budgets. Through sliding windows, exact-capacity controls, adapter interventions, and a scale study, we characterize intermediate depth as a useful search prior and test how well its objective advantage carries over to behavioral outcomes.

Our setup also differs from sparse-update methods that choose a new random or learned subset during optimization. Each allocation in our study is fixed before training, uses the same DPO data and optimizer recipe, and produces an independently evaluable adapter. Contiguous windows preserve the network's depth order and make location interpretable through a single coordinate. The all-depth Matched control then separates the value of concentration in one window from the value of merely reducing trainable capacity. This design brings together parameter-efficient tuning, layer-wise analysis, and preference evaluation in a single controlled setting.

\section{Methodology}
\label{sec:methodology}

\subsection{Layer-Allocation Hypotheses}

We use decoder depth as the coordinate for allocating a fixed trainable budget.

\begin{definition}[Depth Partition]
Let a Transformer model contain $N$ ordered layers, $\mathcal{T} = \{L_1, L_2, \ldots, L_N\}$. A \textbf{depth partition} divides these layers into $K$ disjoint contiguous blocks, $\Pi = \{\mathcal{S}_1, \mathcal{S}_2, \ldots, \mathcal{S}_K\}$, with $\mathcal{S}_k = \{L_i : i \in I_k\}$. The blocks preserve the model's computational order, $F(\mathbf{x}; \Theta) = f_K \circ f_{K-1} \circ \cdots \circ f_1(\mathbf{x})$.
\end{definition}

We test two empirical hypotheses.

\begin{hypothesis}[Depth-Sensitive Allocation]
\label{hyp:specialization}
For a fixed LoRA parameter budget and training recipe, the DPO objective varies with the depth of a contiguous target window. A coarse depth partition can therefore narrow the search space for finer window selection. We test this in Sections~\ref{subsec:block_sweep}--\ref{subsec:matched_capacity} with three training seeds and exact-capacity controls.
\end{hypothesis}

\begin{hypothesis}[Objective--Behavior Relationship]
\label{hyp:correspondence}
For a preference loss $\ell_m$, optimization activity may concentrate in particular depth ranges. We quantify this activity as
$$
g_k(\ell_m) := \mathbb{E}_{\mathbf{x} \sim \mathcal{D}} \left\| \frac{\partial \ell_m(\mathbf{x})}{\partial \Theta_k} \right\|
$$
and ask whether objective differences across depth carry over to generation preferences. Because objective and behavioral rankings can diverge, we evaluate them separately. Section~\ref{sec:mechanism} reports trained-adapter interventions alongside gradient RMS, LoRA update norms, and representation drift.
\end{hypothesis}

\subsection{Implementation}

\subsubsection{Positional Block Partitioning}

Following Hypothesis~\ref{hyp:specialization}, we define $\Pi$ using three positional blocks:
\begin{itemize}
    \item \textbf{Local Block ($\mathcal{S}_{\text{local}}$):} the initial one-third of layers.
    \item \textbf{Intermediate Block ($\mathcal{S}_{\text{mid}}$):} the middle one-third of layers.
    \item \textbf{Global Block ($\mathcal{S}_{\text{global}}$):} the final one-third of layers.
\end{itemize}
These names are positional shorthand; they do not imply that each block serves an exclusive functional role. A deterministic rule distributes $N$ layers as evenly as possible across the three blocks (exact indices in Appendix~\ref{sec:app_layer_partition}). Section~\ref{subsec:block_sweep} refines the coarse partition with overlapping windows.

\subsubsection{Alignment Objective: Direct Preference Optimization (DPO)}

We adopt the DPO loss as our alignment objective. 
For a preference tuple $(x, y_w, y_l)$ where response $y_w$ 
is preferred over $y_l$ for prompt $x$, the loss is defined as:
\begin{align}
\mathcal{L}_{\text{DPO}} 
= & -\mathbb{E} \Bigg[
  \log \sigma \Bigg(
     \beta \log 
     \frac{\pi_\theta(y_w|x)}{\pi_{\text{ref}}(y_w|x)}
     \notag \\
  & \qquad\qquad\qquad
     -\,
     \beta \log 
     \frac{\pi_\theta(y_l|x)}{\pi_{\text{ref}}(y_l|x)}
  \Bigg)
\Bigg]
\label{eq:dpo_loss}
\end{align}
where $\pi_\theta$ is the policy model being optimized,
$\pi_{\text{ref}}$ is a frozen reference model, and 
$\beta$ is a temperature parameter.

\subsubsection{Targeted Updates via Low-Rank Adaptation (LoRA)}

We use LoRA to confine parameter updates to a chosen block $\mathcal{S}_k$, treating it as a \textbf{subspace selector} for the depth-sensitive allocation in Hypothesis~\ref{hyp:specialization}.

We freeze the base model and inject trainable low-rank matrices into the self-attention projections of layers in $\mathcal{S}_k$ only, leaving all FFN weights frozen. Optimization updates the LoRA parameters $\Theta_{k, \text{LoRA}}$:
\begin{equation}
\Theta_{k, \text{LoRA}}^{(t+1)} \leftarrow \Theta_{k, \text{LoRA}}^{(t)} - \eta \nabla_{\Theta_{k, \text{LoRA}}} \mathcal{L}_{\text{DPO}}.
\label{eq:lora_update}
\end{equation}

\subsubsection{Fixed-Budget Allocations}

An allocation specifies a set of adapted layers $I$ and a rank $r_i$ for each $i\in I$. All experiments target the $q$, $k$, $v$, and $o$ attention projections and keep $\alpha_i/r_i=2$. \textbf{Middle} denotes the allocation that adapts the Intermediate block at rank 16. \textbf{Selected} assigns rank 16 to the representative window returned by the procedure below. \textbf{All} assigns rank 16 to every layer and serves as the conventional higher-capacity reference.

\textbf{Matched} spreads Middle's aggregate rank across every layer. For Qwen2.5-7B, Middle adapts nine layers and has rank-sum $9\times16=144$; Matched uses rank 5 in 24 layers and rank 6 in layers 3, 10, 17, and 24. For Llama-3.1-8B, Middle adapts eleven layers and has rank-sum $11\times16=176$; Matched alternates ranks 5 and 6 across the 32 layers. These constructions yield exactly 3.244M and 4.686M trainable parameters for both members of each Middle--Matched pair. Automated checks verify the targeted projections, layer indices, and parameter counts before each training run.

\subsubsection{Coarse-to-Fine Window Selection}

The depth thirds provide a model-independent coarse partition. We refine the middle candidate with overlapping windows whose width equals the model's middle-third width. Every window is trained with seeds 13, 42, and 77 and ranked by mean teacher-forced dev DPO loss. The window with the lowest mean loss serves as the basin reference. Window $i$ belongs to the low-loss basin if its 95\% seed-$t$ interval overlaps the reference interval or if $|\bar\ell_i-\bar\ell_{\min}|<s_{\mathrm{pool}}$, where $s_{\mathrm{pool}}=\sqrt{(s_i^2+s_{\min}^2)/2}$. Among basin members, the frozen representative is the one whose center is closest to the fixed middle-third center, with lower mean loss as a tiebreaker. Preference accuracy, implicit margin, and chosen-token KL provide complementary diagnostics but do not override the loss criterion. Generation results play no role in window selection and are computed only after the representative has been frozen.

This rule identifies a stable region despite seed variability and gives the fixed depth-thirds prior a concrete operational role: Middle is the default candidate, and the window scan provides a lightweight adjustment when neighboring windows fall in the same loss basin. Section~\ref{subsec:block_sweep} reports the full three-seed results, and Appendix~\ref{sec:app_objective} lists every window.

\section{Experiments}
\label{sec:Experiment}

\subsection{Experimental Setup and Evaluation Metrics}
\label{subsec:setup_metrics}

\paragraph{Models and training.} The controlled study uses Qwen2.5-7B \citep{qwen2024qwen25} and Llama-3.1-8B \citep{grattafiori2024llama3} as reference models and adds Qwen2.5-3B/14B for scale. Models train for one epoch on 10{,}000 Anthropic HH-RLHF pairs \citep{bai2022training} with learning rate $10^{-5}$, $\beta=0.1$, effective batch size 16, and bfloat16 precision. LoRA adapts the $\{q,k,v,o\}$ attention projections ($r=16$, $\alpha=32$, dropout $0.05$). Maximum prompt and total sequence lengths are 1{,}024 and 2{,}048 tokens; training seeds are 13, 42, and 77. Generation uses greedy decoding from the seed-42 checkpoint with a 512-token cap. Appendix~\ref{sec:app_hyperparams} records the complete software and training configuration.

\paragraph{Controlled-pipeline integrity.} All headline results use a corrected data pipeline that preserves exact prompt--completion boundaries under the corresponding tokenizer (Appendix~\ref{sec:app_judge_prompt}). Earlier single-seed artifacts from a prior pipeline remain in the exploratory appendix for provenance but do not enter the controlled comparisons.

\paragraph{Splits and uncertainty.} Frozen, disjoint subsets contain 10{,}000 training pairs, 300 dev prompts for teacher-forced DPO loss and window selection, 1{,}000 main-test prompts for generation, and 180 human-pool prompts. Row indices are unique within each split, and prompt text is verified before paired evaluation. Objective effects use 95\% seed-$t$ intervals over the three training seeds. Generation effects use paired prompt bootstrap intervals; human intervals resample prompts so that the three votes for one prompt remain clustered. An interval containing zero is reported as unresolved. Split-integrity checks and valid-count accounting appear in Appendix~\ref{sec:app_protocol}.

\paragraph{Teacher-forced evaluation.} On the frozen 300-prompt dev set, we compute DPO loss, preference accuracy, implicit margin, and chosen-token $\mathrm{KL}(\pi_\theta\Vert\pi_{\mathrm{ref}})$ from each loaded checkpoint. DPO loss is the selection criterion because it directly matches the training objective; the remaining metrics diagnose whether a lower loss coincides with consistent preference movement. For each checkpoint, the policy pass enables the adapter while the reference pass disables it, keeping tokenization and base weights fixed.

\paragraph{Behavioral evaluation.} Qwen3-Max scores 16 dimensions on a 1--5 scale, grouped into language quality, logic and coherence, content quality, safety, task execution, and meta-level utility. DeepSeek-v4-pro and GPT-5.5 provide pairwise judgments with mandatory A/B order swapping. IFEval \citep{zhou2023instruction}, AlpacaEval~2 Length-Controlled \citep{dubois2025length}, and TruthfulQA \citep{lin2021truthfulqa} add programmatic and auxiliary checks. Empty, truncated, and failed generations remain in the denominators of all reported rates. Full endpoint and prompt details appear in Appendix~\ref{sec:app_judge_prompt}.

\subsection{Multi-Seed Window Search}
\label{subsec:block_sweep}

We train eight overlapping windows per reference model at a fixed parameter budget: nine layers for Qwen2.5-7B and eleven for Llama-3.1-8B. Qwen starts are $\{0,3,6,9,10,13,16,19\}$ and Llama starts are $\{0,3,6,9,11,14,17,21\}$. Every window uses the same data order, optimizer, rank, projections, and three training seeds. Applying the selection rule from \S\ref{sec:methodology}, the procedure selects Qwen layers 10--18 and Llama layers 9--19 as frozen representatives; the minimum-mean windows are 13--21 and 6--16, respectively.

\begin{figure*}[t]
    \centering
    \includegraphics[width=\linewidth]{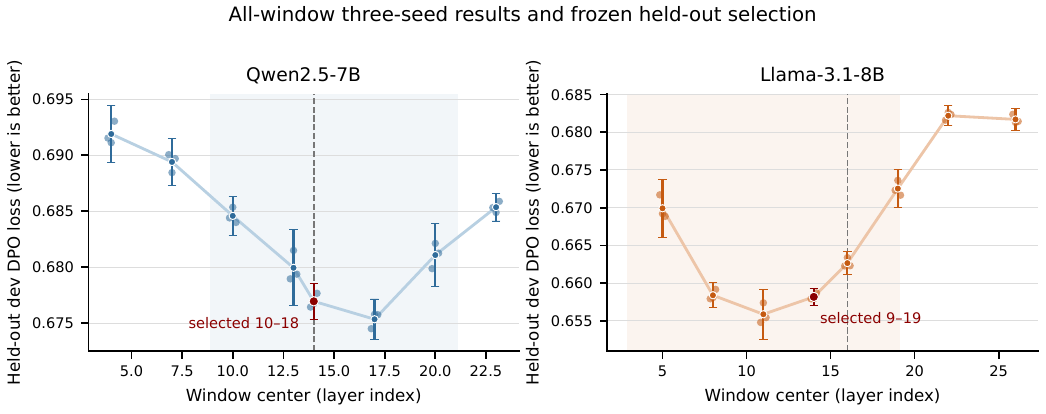}
    \caption{Held-out DPO loss for every contiguous window across seeds 13, 42, and 77. Small points are seed values; large points and bars show means and 95\% seed-$t$ intervals. Gray dashed lines mark the middle-third center; red points mark the frozen representative windows. Qwen's low-loss basin contains 9--17, 10--18, and 13--21; Llama's contains 3--13, 6--16, and 9--19.}
    \label{fig:window_multiseed}
    \label{fig:uwave}
\end{figure*}

Both models exhibit a broad low-loss basin around intermediate depth, though the exact minimum varies. Qwen window means drop from 0.6919 (layers 0--8) to a minimum of 0.6753 (layers 13--21) before rising again; Llama's minimum is 0.6559 (layers 6--16), with the two deepest windows rising above 0.681. Qwen's basin contains windows 9--17, 10--18, and 13--21; Llama's contains 3--13, 6--16, and 9--19. The center-nearest rule therefore returns the fixed Middle for Qwen and a middle-adjacent window for Llama. Across seeds, the result supports a region-level prior with model-specific refinement.

Preference accuracy, implicit margin, and chosen-token KL provide complementary consistency checks but do not change the selected representative. Crucially, the selector was frozen before any generation took place: the behavioral test set supplies no search signal, and each judge evaluates a single predeclared checkpoint per model.

\subsection{Matched-Capacity Objective Control}
\label{subsec:matched_capacity}

Matched allocates an aggregate rank of 144 across all 28 Qwen layers and 176 across all 32 Llama layers, exactly matching Middle's 3.244M and 4.686M trainable parameters. Rank and alpha vary together so that $\alpha/r=2$ in every adapted layer. The Qwen pattern uses rank 6 at four evenly spaced layers and rank 5 elsewhere; the Llama pattern alternates ranks 5 and 6. The result is an all-depth adapter with exactly the same trainable budget as Middle.

Middle has lower dev loss than Matched on Qwen ($\Delta=-0.0086$, 95\% CI $[-0.0095,-0.0078]$) and Llama ($-0.0057$, $[-0.0109,-0.0006]$). The interval excludes zero for both models, establishing a placement effect at fixed capacity. Selected versus the larger All configuration is unresolved on Qwen ($+0.0028$, $[-0.0010,+0.0066]$) and Llama ($+0.0036$, $[-0.0024,+0.0096]$). Concentrating capacity in the middle region thus improves the objective over an equal-capacity all-depth spread. The Selected--All gap remains unresolved with three seeds, even though All uses roughly three times as many trainable parameters.

\begin{figure}[t]
    \centering
    \includegraphics[width=\linewidth]{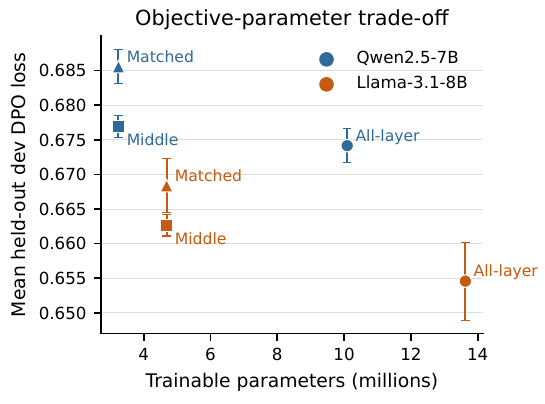}
    \caption{Mean held-out DPO loss versus trainable parameters; bars are 95\% seed-$t$ intervals. Middle improves on the exact-capacity all-depth Matched control for both models; All is the higher-capacity all-depth reference.}
    \label{fig:matched_pareto}
\end{figure}

\subsection{Behavioral Validation}
\label{subsec:eval_validation}

Table~\ref{tab:behavior_summary} reports behavioral effects alongside the objective results. On Qwen2.5-7B, Selected improves over Base by $+0.152$ on the Qwen3-Max Overall scale ($[+0.093,+0.212]$). Against All, Selected is unresolved across all four evaluators: Qwen3-Max ($-0.020$), DeepSeek-v4-pro ($-0.043$), GPT-5.5 ($-0.029$), and the human panel. In this setting, the behavioral gap between Selected and All is small relative to prompt-level uncertainty.

Llama tells a different story. DeepSeek-v4-pro and GPT-5.5 both favor All over Selected by 6.1 and 7.7 net-preference points, with intervals excluding zero; the human point estimate agrees, though its wider interval includes zero. Against the equal-capacity Matched control, Qwen Selected is favored by DeepSeek ($+0.081$) but unresolved under GPT-5.5 and Qwen3-Max; Llama Middle is disfavored against Matched by both judges. IFEval differences are mostly within two percentage points; 27 of 28 intervals include zero, with one resolved Qwen2.5-14B Middle--Matched result on loose instruction accuracy (Appendix~\ref{sec:app_eval_validation}). AlpacaEval~2 assigns All a descriptively higher length-controlled win rate on both models (Qwen 53.98 vs.\ 50.00; Llama 4.99 vs.\ 4.28), but the Qwen estimator places several conditions at its 50.00 regularization floor, so we draw no inferential conclusion.

\begin{table*}[t]
\centering\small
\caption{Representative controlled behavioral effects. Effects are net preference for the first allocation, except Qwen3-Max (difference on its 1--5 Overall score). $\star$ marks a 95\% interval excluding zero. Complete judge and benchmark tables are in Appendix~\ref{sec:app_eval_validation}.}
\label{tab:behavior_summary}
\setlength{\tabcolsep}{4pt}
\begin{tabular*}{0.90\textwidth}{@{\extracolsep{\fill}}llllr@{}}
\toprule
\textbf{Model} & \textbf{Comparison} & \textbf{Evaluator} & \textbf{Effect [95\% CI]} & \textbf{$n$} \\
\midrule
Qwen2.5-7B & Selected $-$ All & Qwen3-Max & $-0.020\;[-0.073,+0.031]$ & 1000 \\
Qwen2.5-7B & Selected $-$ All & DeepSeek-v4-pro & $-0.043\;[-0.087,+0.002]$ & 1000 \\
Qwen2.5-7B & Selected $-$ All & Human & $-0.089\;[-0.203,+0.028]$ & 120 \\
Llama-3.1-8B & Selected $-$ All & DeepSeek-v4-pro & $-0.061\;[-0.110,-0.013]^{\star}$ & 1000 \\
Llama-3.1-8B & Selected $-$ All & Human & $-0.069\;[-0.197,+0.061]$ & 120 \\
Qwen2.5-7B & Selected $-$ Matched & DeepSeek-v4-pro & $+0.081\;[+0.035,+0.126]^{\star}$ & 1000 \\
Llama-3.1-8B & Middle $-$ Matched & DeepSeek-v4-pro & $-0.076\;[-0.124,-0.030]^{\star}$ & 1000 \\
\bottomrule
\end{tabular*}
\end{table*}

\noindent\textbf{Blind human evaluation.}\quad Three independent annotators compare 120 prompts per model, with randomized A/B order, ties allowed, and six repeated quality-control prompts per annotator. Model family, allocation, and response orientation are hidden. Annotators read temperature-zero Chinese translations of each prompt and output; identical source strings always receive the same canonical translation. The comparison therefore measures preference over translated Chinese renderings. Formal estimates exclude the repeated prompts and retain one Llama item for which both outputs are empty as a tie.

Qwen yields 114 Selected wins, 100 ties, and 146 All wins; Llama yields 133 Selected wins, 69 ties, and 158 All wins. The tie-inclusive net preferences are $-0.089$ ($[-0.203,+0.028]$) and $-0.069$ ($[-0.197,+0.061]$), respectively. Fleiss' $\kappa$ is 0.401/0.514 and repeated-item consistency is 0.833/0.889 for Qwen/Llama. Three-way prompt-level agreement---exact agreement between the human-majority A/B/tie label and the LLM judge's aggregated label---is 0.400/0.492 for Qwen and 0.583/0.658 for Llama under DeepSeek/GPT-5.5. The higher Llama agreement is consistent with the shared direction toward All across evaluators; on Qwen, evaluator judgments are more dispersed.

\begin{figure*}[t]
    \centering
    \includegraphics[width=\linewidth]{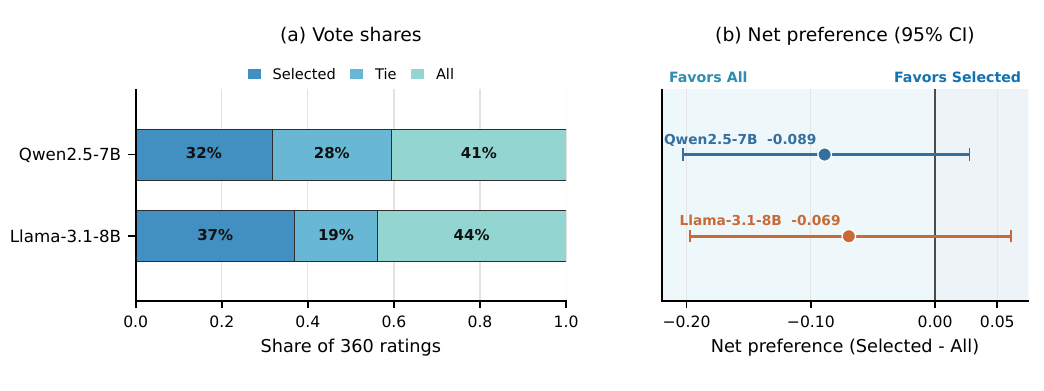}
    \caption{Blinded human comparison of the frozen Selected window against All. Left: fractions of 360 formal votes per model. Right: tie-inclusive net preference with 95\% prompt-clustered bootstrap intervals. Both intervals include zero.}
    \label{fig:human_eval}
\end{figure*}

\subsection{Scale and Efficiency}
\label{subsec:robustness_xdata}

\paragraph{Qwen scale study.} The same deterministic thirds give Middle layers 12--23 at 3B, 10--18 at 7B, and 16--31 at 14B. At 3B, dev loss favors All over Middle ($\Delta_{\text{Mid-All}}=+0.0072$, 95\% CI $[+0.0035,+0.0109]$); at 14B the difference is unresolved ($+0.0014$, $[-0.0058,+0.0085]$). Qwen3-Max behavior is also unresolved between Middle and All at both scales: $-0.005$ $[-0.063,+0.053]$ at 3B and $+0.001$ $[-0.049,+0.051]$ at 14B. Middle and All each improve over Base at 3B ($+0.201/+0.206$) and 14B ($+0.429/+0.428$).

The 14B exact-capacity comparison adds a behavioral check on placement. Middle exceeds Matched by $+0.094$ ($[+0.031,+0.156]$), and All exceeds Matched by $+0.093$ ($[+0.033,+0.154]$), on 999 valid prompts. These point estimates are nearly identical, while the direct Middle--All difference remains unresolved. The equal-capacity all-depth distribution falls behind both. Across the three scales, the depth prior remains useful, though the precise objective gap relative to All varies by scale.

\paragraph{Measured efficiency.} Appendix Table~\ref{tab:efficiency_full} reports the seed-42 NVIDIA H800 implementation profile over steps 50--600. Because Middle and Matched have the same trainable parameter count and nearly the same serialized adapter size, the runtime gap between them isolates the effect of placing adapters in a contiguous block versus spreading them across every layer. Middle records 17.6--18.1~GiB lower peak memory, 18.6--19.2\% shorter steps, and 22.8--23.7\% higher throughput. All uses roughly three times the trainable parameters and optimizer-state storage of Middle. Every allocation still executes a full forward pass through the frozen base; the measured savings are specific to this implementation, hardware, and optimizer-state layout. Supplementary HelpSteer2, TruthfulQA, SimPO, and layer-wise diagnostics serve as exploratory evidence outside the headline comparisons.

\section{Trained-Adapter Interventions}
\label{sec:mechanism}

We intervene on the trained seed-42 All checkpoint by masking adapter contributions in selected depth blocks while keeping all other weights fixed. A \emph{removal} masks one block's adapters; a \emph{retention} keeps only one block and masks the other two. Both operations are post-training, so they measure how much the trained model depends on adapter contributions from each depth region.

Removing Middle increases dev DPO loss on Qwen by $+0.0124$ ($[+0.0050,+0.0194]$) and on Llama by $+0.0157$ ($[+0.0032,+0.0278]$); removing Llama's selected 9--19 window increases it by $+0.0207$ ($[+0.0044,+0.0364]$). Among the three depth thirds, retaining only Middle preserves the largest fraction of All's objective gain over Base: 0.765 for Qwen and 0.645 for Llama (Shallow: 0.157/0.497; Deep: 0.291/0.127). Retaining Llama's frozen Selected window (layers 9--19) preserves an even larger fraction, 0.801.

\begin{figure*}[t]
    \centering
    \includegraphics[width=\linewidth]{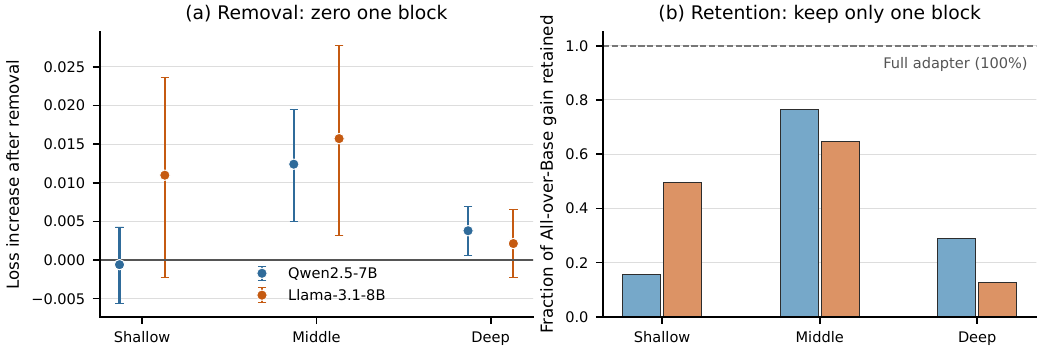}
    \caption{Trained-adapter interventions on the seed-42 All checkpoint. Within each panel, higher values indicate larger measured intervention effects. Left: dev-loss increase after removing one fixed depth-third block, with paired-prompt 95\% intervals. Right: fraction of the All-over-Base dev-loss gain retained when only one fixed depth-third block remains. The Llama Selected 9--19 intervention is reported in the text and Appendix~\ref{sec:app_interventions}.}
    \label{fig:intervention}
\end{figure*}

Generation-level effects are less consistent. For middle-block removal, the Qwen3-Max Overall change is unresolved on both models. On Llama, however, removing the Selected 9--19 adapters reduces Overall by $-0.258$ ($[-0.395,-0.120]$), and removing the shallow block reduces it by $-0.455$ ($[-0.605,-0.311]$). Teacher-forced effects thus peak at middle or middle-adjacent depth, while removing Llama's shallow block also lowers generation quality. Gradient RMS, update norms, and representation drift provide complementary correlational depth profiles (Appendix~\ref{sec:app_mechanism}).

\section{Discussion and Conclusion}
\label{sec:discussion}

\paragraph{What the objective results establish.} The multi-seed windows show that held-out DPO loss changes systematically with allocation depth. Their minima differ across Qwen and Llama, yet both low-loss basins cover middle or middle-adjacent windows. The matched-capacity controls strengthen this result: concentrating the same trainable parameter count in Middle improves the objective over distributing it across every layer. The Selected--All loss differences remain unresolved with three seeds, which puts allocation and capacity on the same practical scale but does not imply that the two strategies are equivalent.

\paragraph{What behavioral evaluation adds.} A lower teacher-forced loss is a useful training signal and selection criterion, but generated responses introduce additional sources of variation: decoding strategy, prompt content, evaluator choice, and rubric design. Qwen Selected versus All remains unresolved across four evaluator views. Llama's external judges favor All, with human votes pointing in the same direction at lower precision. At 3B and 14B, Middle and All each improve over Base and remain unresolved against each other under Qwen3-Max. These outcomes make behavioral validation an explicit second stage of the search procedure.

\paragraph{Interpreting depth.} The positional labels provide a stable coordinate shared across architectures. Adapter removal and retention establish that the location of trained components affects the DPO objective, and selected Llama removals also affect generation. Gradient RMS, update norms, and representation drift show related depth structure, with a secondary late-layer update cluster in Llama. Together, these measurements show where preference-learning signals concentrate in the studied checkpoints and help explain why the same fixed window need not work across the models studied.

\paragraph{Practical workflow.} A resource-constrained practitioner should first check whether all-layer DPO improves over Base on the target generation rubric. If it does, rank-16 Middle LoRA-DPO is a natural starting point. A small set of overlapping neighboring windows can then be trained on the same budget and screened for a low-loss basin using held-out preference data across seeds. The selected adapter should finally be compared with Base and the all-layer reference on the target generation rubric. An exact-capacity all-depth control is especially informative when the goal is to attribute gains to placement. This sequence keeps the layer search small while preserving a direct behavioral decision for deployment.

\paragraph{Conclusion.} Hierarchical Alignment frames preference tuning as a layer-allocation problem. Multi-seed windows, exact-capacity controls, interventions, and scale results show that allocation changes the objective and can alter generated behavior. Intermediate depth provides a useful starting point for the search; the final window should be confirmed by behavioral evaluation on the target task.

\section{Limitations}
\label{sec:limitations}

Our controlled experiments cover decoder-only Qwen models at 3B, 7B, and 14B and Llama at 8B. The main training corpus is HH-RLHF; HelpSteer2-Preference \citep{wang2024helpsteer2preference} serves as a smaller transfer check. A Qwen3-8B case from an earlier pipeline shows no detectable All-versus-Base behavioral headroom and is retained only as a boundary diagnosis, not as evidence about layer ordering (Appendix~\ref{sec:app_qwen3_headroom}). Broader architectures, model families, preference domains, and languages may yield different depth profiles.

The objective analysis uses three training seeds, whereas generation and most judge evaluations use the seed-42 checkpoints; prompt-level bootstrap intervals cover prompt uncertainty while leaving training-seed uncertainty unmeasured. Our frozen window selector uses held-out DPO loss, and objective and behavioral rankings sometimes diverge.

Qwen3-Max supplies the multidimensional score, while DeepSeek-v4-pro and GPT-5.5 provide pairwise checks. These proprietary judges limit reproducibility and may share unknown biases. The blinded human study covers 120 prompts per model comparison with three annotators; its intervals include zero and agreement is moderate. Because annotators judged canonical Chinese translations, these preferences cannot be equated directly with the English automated evaluations, and translation may alter nuance.

We do not compare against learned masks, gradient-selected subsets, or random layer subsets under the same LoRA-DPO budget. The Matched control isolates the effect of concentrating capacity versus spreading it across all layers, but does not test learned layer selection. The adapter-removal experiments establish dependence on trained components, while nonlinear interactions between adapter blocks limit functional localization claims. Efficiency measurements come from a single implementation and hardware environment. Middle uses fewer LoRA parameters than All and shows lower peak memory and shorter steps than the exact-capacity Matched control, but every model still runs a full forward pass through the frozen base, so a general systems-level speedup remains unestablished.

\section{Ethical Considerations}

HH-RLHF and HelpSteer2 may contain biases that preference tuning can amplify, and layer-selective tuning carries dual-use risks. We therefore report uncertainty and evaluator disagreement throughout, and pair proprietary judges with programmatic checks and blinded human evaluation.

All three authors, who are adults and native speakers of Chinese, served as annotators. No external participants were recruited or paid. Model identity, allocation, and response orientation were hidden during annotation, and results are reported without attributing labels to individual authors. Because the study involved no external human participants, no separate participant-consent procedure or institutional ethics review was sought.

\section*{Use of AI Assistants}
AI assistants supported editing and consistency checks. Temperature-zero \texttt{deepseek-chat} produced the Chinese translations used in the human study but did not generate any labels (Appendix~\ref{sec:app_human}). The authors retain full responsibility for the research.

\bibliography{camera_ready}

\appendix
\input{camera_ready_appendix}
\end{document}

%% file: camera_ready_appendix.tex
\captionsetup[table]{skip=4pt}
\setlength{\dbltextfloatsep}{10pt plus 2pt minus 2pt}
\setlength{\dblfloatsep}{8pt plus 2pt minus 2pt}
\renewcommand{\arraystretch}{1.05}

\section{Data Integrity and Split Roles}
\label{sec:app_protocol}

\subsection{Data Roles and Integrity}

The controlled study follows a protocol frozen before the formal run matrix. The training set contains 10{,}000 HH-RLHF preference pairs. Three mutually disjoint subsets drawn from the HH-RLHF test pool have fixed roles: 300 prompts for teacher-forced development metrics and window selection, 1{,}000 prompts for generation-based evaluation, and 180 prompts reserved for human evaluation. The stored row order is fixed. Every original row index is unique within its split, and disjointness is verified using both source row IDs and prompt text before paired evaluation.

\section{Generation and Judge Protocols}
\label{sec:app_judge_prompt}

\subsection{Judges, Endpoints, and Rubrics}

All API evaluations were completed and audited on July 14--15, 2026 UTC. The absolute rubric uses the model identifier \texttt{qwen3-max} through Aliyun's OpenAI-compatible endpoint. Pairwise evaluation uses \texttt{deepseek-v4-pro} through the official DeepSeek endpoint and \texttt{gpt-5.5} through the ZYAI OpenAI-compatible relay. Each pairwise comparison uses one provider throughout, verifies prompt identity, presents both A/B orders, and retains ties. The final pairwise matrix has 1{,}000 valid prompts per row. Incomplete files and failed API rows are excluded; rows from different providers are never merged. No credential is stored with the results.

The 16 Qwen3-Max dimensions are Grammar/Fluency, Token Robustness, Conciseness, Syntactic Complexity, Coherence/Logic, Topic Coherence, Tone, Factuality, Knowledge Depth, Cross-Domain Generalization, Creativity, Safety, Ethics Sensitivity, Meta-Reflection, Instruction Following, and Overall. Each dimension is scored from 1 to 5. The system message defines the model as a professional text-quality evaluator. The generated user template inserts the definitions of all 16 dimensions, five anchored score levels, a fixed score-3 example, the evaluated prompt and response, and an output instruction requiring a JSON array of \texttt{dimension}, integer \texttt{score}, and short \texttt{reason} objects. Validity checks require all 16 dimensions and an integer score in range.

For pairwise evaluation, the system message is: ``You are a fair and expert evaluator comparing two assistant responses. You pay careful attention to helpfulness, correctness, instruction following, and safety. You never favor longer or more verbose answers if they are not clearly more useful.'' The user template inserts the user prompt and responses A/B, asks which response better answers the request without factual or logical errors, is more useful, safe, and appropriate, and requires only \texttt{\{"winner": "A" | "B" | "tie", "reason": "<1 short sentence>"\}}. Every prompt is judged again with the response order swapped; a pair is valid only when both calls parse. The reverse-order label is mapped back to the original A/B frame. If the two calls agree, their shared label is the prompt label; if they disagree, the prompt is recorded as a tie. Thus each of the $n$ prompts contributes one aggregated A/B/tie label, net preference is $(A\text{ wins}-B\text{ wins})/n$, and bootstrap resampling operates at the prompt level.

\subsection{Complete Qwen3-Max and Pairwise Effects}

\begin{table*}[t!]
\centering\footnotesize
\caption{Complete controlled Qwen3-Max Overall effects. Effects are A minus B on the 1--5 scale.}
\label{tab:qwen_rubric_full}
\begin{tabular*}{0.84\textwidth}{@{\extracolsep{\fill}}llrr@{}}
\toprule
\textbf{Comparison (A$-$B)} & \textbf{Effect [95\% CI]} & \textbf{$n$} & \textbf{Interpretation} \\
\midrule
Qwen2.5-7B Selected$-$Base & $+0.152\;[+0.093,+0.212]$ & 1000 & A favored \\
Qwen2.5-7B Selected$-$All & $-0.020\;[-0.073,+0.031]$ & 1000 & unresolved \\
Qwen2.5-7B Selected$-$Matched & $+0.036\;[-0.018,+0.090]$ & 1000 & unresolved \\
Qwen2.5-7B All$-$Matched & $+0.056\;[-0.001,+0.113]$ & 1000 & unresolved \\
Qwen2.5-3B Middle$-$Base & $+0.201\;[+0.135,+0.266]$ & 1000 & A favored \\
Qwen2.5-3B All$-$Base & $+0.206\;[+0.137,+0.273]$ & 1000 & A favored \\
Qwen2.5-3B Middle$-$All & $-0.005\;[-0.063,+0.053]$ & 1000 & unresolved \\
Qwen2.5-14B Middle$-$Base & $+0.429\;[+0.355,+0.505]$ & 1000 & A favored \\
Qwen2.5-14B All$-$Base & $+0.428\;[+0.353,+0.503]$ & 1000 & A favored \\
Qwen2.5-14B Middle$-$All & $+0.001\;[-0.049,+0.051]$ & 1000 & unresolved \\
Qwen2.5-14B Matched$-$Base & $+0.335\;[+0.265,+0.404]$ & 999 & A favored \\
Qwen2.5-14B Middle$-$Matched & $+0.094\;[+0.031,+0.156]$ & 999 & A favored \\
Qwen2.5-14B All$-$Matched & $+0.093\;[+0.033,+0.154]$ & 999 & A favored \\
\bottomrule
\end{tabular*}
\end{table*}

\begin{table*}[t!]
\centering\footnotesize
\caption{Complete controlled order-swapped pairwise effects. Effects are tie-inclusive net preference for A over B. Each row has 1{,}000 valid prompt pairs.}
\label{tab:pairwise_full}
\begin{tabular*}{\textwidth}{@{\extracolsep{\fill}}lllrl@{}}
\toprule
\textbf{Comparison (A$-$B)} & \textbf{Judge} & \textbf{Effect [95\% CI]} & \textbf{$n$} & \textbf{Interpretation} \\
\midrule
Qwen2.5-7B Selected$-$All & DeepSeek-v4-pro & $-0.043\;[-0.087,+0.002]$ & 1000 & unresolved \\
 & GPT-5.5 & $-0.029\;[-0.079,+0.021]$ & 1000 & unresolved \\
Qwen2.5-7B Selected$-$Matched & DeepSeek-v4-pro & $+0.081\;[+0.035,+0.126]$ & 1000 & A favored \\
 & GPT-5.5 & $+0.049\;[-0.002,+0.100]$ & 1000 & unresolved \\
Llama-3.1-8B Selected$-$All & DeepSeek-v4-pro & $-0.061\;[-0.110,-0.013]$ & 1000 & B favored \\
 & GPT-5.5 & $-0.077\;[-0.130,-0.023]$ & 1000 & B favored \\
Llama-3.1-8B Middle$-$Matched & DeepSeek-v4-pro & $-0.076\;[-0.124,-0.030]$ & 1000 & B favored \\
 & GPT-5.5 & $-0.083\;[-0.136,-0.030]$ & 1000 & B favored \\
Qwen2.5-14B Middle$-$All & DeepSeek-v4-pro & $+0.018\;[-0.027,+0.062]$ & 1000 & unresolved \\
 & GPT-5.5 & $+0.042\;[-0.008,+0.091]$ & 1000 & unresolved \\
\bottomrule
\end{tabular*}
\end{table*}

\subsection{Programmatic Benchmarks}
\label{sec:app_eval_validation}

IFEval uses 541 prompts containing 834 separately scored instructions. Table~\ref{tab:ifeval_full} reports all controlled allocation comparisons. Effects are percentage-point differences for A minus B. Twenty-seven of 28 intervals contain zero; the one resolved result is Qwen2.5-14B Middle over Matched on loose instruction accuracy.

\begin{table*}[t!]
\centering\footnotesize
\setlength{\tabcolsep}{2pt}
\caption{Complete IFEval effects in percentage points. SP/SI are strict prompt/instruction accuracy; LP/LI are loose prompt/instruction accuracy.}
\label{tab:ifeval_full}
\begin{adjustbox}{max width=\textwidth}
\begin{tabular}{lrrrr}
\toprule
\textbf{Comparison (A$-$B)} & \textbf{SP [95\% CI]} & \textbf{SI [95\% CI]} & \textbf{LP [95\% CI]} & \textbf{LI [95\% CI]} \\
\midrule
Qwen2.5-7B Selected$-$All & $-0.92\;[-2.96,+1.11]$ & $-1.80\;[-3.89,+0.35]$ & $-0.37\;[-2.77,+2.03]$ & $-1.80\;[-4.19,+0.50]$ \\
Qwen2.5-7B Selected$-$Matched & $-1.29\;[-3.51,+0.92]$ & $-0.72\;[-2.81,+1.34]$ & $-0.74\;[-3.33,+1.85]$ & $-0.72\;[-3.06,+1.67]$ \\
Llama-3.1-8B Selected$-$All & $-0.92\;[-2.59,+0.74]$ & $-0.72\;[-2.00,+0.59]$ & $-1.11\;[-2.77,+0.55]$ & $-0.84\;[-2.14,+0.48]$ \\
Llama-3.1-8B Middle$-$Matched & $-0.37\;[-1.85,+1.11]$ & $-0.24\;[-1.53,+0.97]$ & $-0.55\;[-2.22,+0.92]$ & $-0.24\;[-1.59,+1.08]$ \\
Qwen2.5-3B Middle$-$All & $+0.18\;[-1.85,+2.22]$ & $-0.72\;[-2.44,+0.97]$ & $+0.55\;[-1.48,+2.77]$ & $-0.48\;[-2.32,+1.35]$ \\
Qwen2.5-14B Middle$-$All & $-0.18\;[-2.03,+1.66]$ & $-0.36\;[-2.06,+1.32]$ & $+0.37\;[-1.85,+2.59]$ & $+0.36\;[-1.58,+2.36]$ \\
Qwen2.5-14B Middle$-$Matched & $+0.92\;[-1.66,+3.51]$ & $+1.68\;[-0.60,+3.98]$ & $+1.66\;[-1.11,+4.44]$ & $+2.76\;[+0.37,+5.16]$ \\
\bottomrule
\end{tabular}
\end{adjustbox}
\end{table*}

AlpacaEval~2 LC uses the \nolinkurl{weighted_alpaca_eval_gpt-4o-mini-2024-07-18} configuration, with 805 prompts per condition. The LC estimator can regularize near-reference raw win rates to exactly 50.00; this value is not an equivalence test.

\begin{table*}[t!]
\centering\footnotesize
\caption{Controlled AlpacaEval~2 results. Empty counts include empty or immediate-EOS generations.}
\label{tab:alpaca_fresh}
\begin{tabular*}{0.84\textwidth}{@{\extracolsep{\fill}}llrrrr@{}}
\toprule
\textbf{Model} & \textbf{Allocation} & \textbf{Empty} & \textbf{Raw WR $\pm$ SE} & \textbf{LC WR $\pm$ SE} & \textbf{Avg. chars} \\
\midrule
Qwen2.5-7B & Base & 20 & $52.22\pm1.72$ & $50.00\pm0.00$ & 2893 \\
 & Matched & 19 & $51.77\pm1.72$ & $50.00\pm0.00$ & 2943 \\
 & Selected & 19 & $52.45\pm1.72$ & $50.00\pm0.00$ & 2720 \\
 & All & 13 & $54.16\pm1.72$ & $53.98\pm0.02$ & 2801 \\
Llama-3.1-8B & Base & 67 & $5.81\pm0.81$ & $3.44\pm0.07$ & 6223 \\
 & Matched & 58 & $6.03\pm0.82$ & $3.52\pm0.07$ & 6135 \\
 & Middle & 56 & $7.59\pm0.91$ & $4.59\pm0.08$ & 5978 \\
 & Selected & 60 & $7.19\pm0.89$ & $4.28\pm0.07$ & 5902 \\
 & All & 53 & $7.47\pm0.90$ & $4.99\pm0.10$ & 5955 \\
\bottomrule
\end{tabular*}
\end{table*}

For plain-text records, the controlled pipeline treats the final space at the prompt--completion boundary as the first character of each completion. This preserves the exact concatenated dialogue and ensures that the prompt token IDs are a prefix of both tokenized prompt--completion sequences. An audit found a tokenizer-boundary mismatch in legacy adapters; no legacy adapter or generation enters the controlled headline tables.

\section{Training and Allocation Protocol}

\subsection{Training and Decoding}
\label{sec:app_hyperparams}

Table~\ref{tab:full_hyperparams} gives the complete shared recipe. Qwen2.5-14B uses micro-batch 2, accumulation 8, and gradient checkpointing for memory; its effective batch and all optimization settings remain unchanged. The recorded software environment is Python 3.11.15, PyTorch 2.10.0+cu128, Transformers 5.6.1, TRL 1.2.0, PEFT 0.19.1, and Datasets 4.8.4.

\begin{table}[t!]
\centering\footnotesize
\caption{Training and generation settings for controlled HH-RLHF experiments.}
\label{tab:full_hyperparams}
\begin{tabularx}{\linewidth}{Xl}
\toprule
\textbf{Item} & \textbf{Value} \\
\midrule
Training pairs / epochs & 10{,}000 / 1 \\
Learning rate / weight decay & $1\times10^{-5}$ / 0 \\
Optimizer & AdamW ($\beta_1=0.9$, $\beta_2=0.999$) \\
DPO $\beta$ & 0.1 \\
Precision & bfloat16 \\
Micro-batch / accumulation & 4 / 4 \\
Effective batch & 16 \\
LoRA targets & $q,k,v,o$ projections \\
Standard LoRA & $r=16$, $\alpha=32$, dropout 0.05 \\
Prompt / total length & 1{,}024 / 2{,}048 \\
Training seeds & 13, 42, 77 \\
Generation & greedy, seed 42, max 512 new tokens \\
\bottomrule
\end{tabularx}
\end{table}

\subsection{Compute and Artifact Terms}

The recorded controlled run matrix contains 77 completed single-GPU training runs. Four NVIDIA H800 GPUs were used as independent single-GPU workers; summed recorded training time is approximately 22.1 GPU-hours. This training-only total excludes generation and API-based judging.

HH-RLHF is distributed under the MIT License; HelpSteer2-Preference under CC BY 4.0; Qwen2.5 and Qwen3 under Apache 2.0; and Llama 3.1 under the Llama 3.1 Community License. We used these artifacts for research preference tuning and evaluation.

Teacher-forced metrics compare the adapted policy with the unadapted reference on the same preferred and dispreferred completions. Objective intervals are two-sided 95\% Student-$t$ intervals across the three training seeds ($\mathit{df}=2$). Generation uses the seed-42 checkpoint. Pairwise and rubric effects use paired-prompt bootstrap intervals; human intervals resample prompts while keeping each prompt's three votes together.

\subsection{Positional Partitions and Exact-Capacity Allocation}
\label{sec:app_layer_partition}

Layer indices begin at zero. Local, Intermediate, and Global denote contiguous depth ranges only. Table~\ref{tab:layer_partition} lists every controlled partition. The selected Llama window is middle-adjacent and differs from the deterministic middle third.

\begin{table*}[t!]
\centering\footnotesize
\caption{Depth partitions in the controlled study.}
\label{tab:layer_partition}
\begin{tabular*}{0.78\textwidth}{@{\extracolsep{\fill}}lrrrrr@{}}
\toprule
\textbf{Model} & \textbf{Layers} & \textbf{Local} & \textbf{Intermediate} & \textbf{Global} & \textbf{Selected} \\
\midrule
Qwen2.5-3B  & 36 & 0--11 & 12--23 & 24--35 & 12--23 \\
Qwen2.5-7B  & 28 & 0--9  & 10--18 & 19--27 & 10--18 \\
Llama-3.1-8B & 32 & 0--10 & 11--21 & 22--31 & 9--19 \\
Qwen2.5-14B & 48 & 0--15 & 16--31 & 32--47 & 16--31 \\
\bottomrule
\end{tabular*}
\end{table*}

Matched preserves $\alpha/r=2$ in every layer and matches the aggregate rank of Middle. For Qwen2.5-7B, Middle has rank-sum $9\times16=144$; Matched uses rank 5 in 24 layers and rank 6 at layers 3, 10, 17, and 24. For Llama-3.1-8B, Middle has rank-sum $11\times16=176$; Matched uses rank 5 in the 16 even layers and rank 6 in the 16 odd layers. Automated checks verified the targeted projections and the exact trainable-parameter count before each run.

\section{Complete Objective-Level Results}
\label{sec:app_objective}

\subsection{Multi-Seed Window Search}

The frozen selector ranks mean dev DPO loss across seeds 13, 42, and 77. Let the minimum-mean reference have mean $\bar\ell_{\min}$ and sample standard deviation $s_{\min}$. Window $i$ enters the low-loss basin if its 95\% Student-$t$ interval overlaps the reference interval or if $|\bar\ell_i-\bar\ell_{\min}|<\sqrt{(s_i^2+s_{\min}^2)/2}$. Among basin members, the selected representative is the one whose center is closest to the fixed middle-third center, with lower mean loss as a tiebreaker. Preference accuracy, mean implicit margin, and chosen-token KL are consistency checks and do not replace the loss criterion.

\begin{table*}[t!]
\centering\scriptsize
\setlength{\tabcolsep}{3pt}
\caption{Qwen2.5-7B held-out window results across three training seeds.}
\label{tab:qwen_windows_fresh}
\begin{tabular*}{\textwidth}{@{\extracolsep{\fill}}lrrrrr@{}}
\toprule
\textbf{Layers} & \textbf{Dev loss $\pm$ SD} & \textbf{95\% CI} & \textbf{Pref. acc.} & \textbf{Margin} & \textbf{Chosen KL} \\
\midrule
0--8   & $0.691907\pm0.001012$ & $[0.689394,0.694420]$ & $0.5322\pm0.0168$ & $0.00441\pm0.00177$ & $0.00247\pm0.00051$ \\
3--11  & $0.689406\pm0.000844$ & $[0.687311,0.691502]$ & $0.5278\pm0.0096$ & $0.01027\pm0.00161$ & $0.00350\pm0.00094$ \\
6--14  & $0.684585\pm0.000699$ & $[0.682849,0.686320]$ & $0.5544\pm0.0084$ & $0.02136\pm0.00200$ & $0.00486\pm0.00133$ \\
9--17  & $0.679940\pm0.001362$ & $[0.676556,0.683324]$ & $0.5656\pm0.0139$ & $0.03267\pm0.00347$ & $0.00620\pm0.00120$ \\
10--18$^{\dagger}$ & $0.676943\pm0.000640$ & $[0.675354,0.678532]$ & $0.5822\pm0.0069$ & $0.03910\pm0.00057$ & $0.00634\pm0.00108$ \\
13--21$^{\star}$ & $0.675337\pm0.000730$ & $[0.673524,0.677150]$ & $0.6156\pm0.0150$ & $0.04068\pm0.00129$ & $0.00406\pm0.00029$ \\
16--24 & $0.681079\pm0.001137$ & $[0.678254,0.683904]$ & $0.6056\pm0.0255$ & $0.02662\pm0.00249$ & $0.00223\pm0.00024$ \\
19--27 & $0.685363\pm0.000498$ & $[0.684127,0.686599]$ & $0.5900\pm0.0033$ & $0.01664\pm0.00091$ & $0.00111\pm0.00006$ \\
\bottomrule
\multicolumn{6}{l}{$^{\star}$ minimum mean; $^{\dagger}$ frozen representative. Basin: 9--17, 10--18, 13--21.}
\end{tabular*}
\end{table*}

\begin{table*}[t!]
\centering\scriptsize
\setlength{\tabcolsep}{3pt}
\caption{Llama-3.1-8B held-out window results across three training seeds.}
\label{tab:llama_windows_fresh}
\begin{tabular*}{\textwidth}{@{\extracolsep{\fill}}lrrrrr@{}}
\toprule
\textbf{Layers} & \textbf{Dev loss $\pm$ SD} & \textbf{95\% CI} & \textbf{Pref. acc.} & \textbf{Margin} & \textbf{Chosen KL} \\
\midrule
0--10  & $0.669917\pm0.001556$ & $[0.666051,0.673782]$ & $0.5800\pm0.0033$ & $0.09603\pm0.01146$ & $0.05121\pm0.01632$ \\
3--13  & $0.658393\pm0.000675$ & $[0.656716,0.660070]$ & $0.5922\pm0.0038$ & $0.12261\pm0.00336$ & $0.04683\pm0.00266$ \\
6--16$^{\star}$ & $0.655876\pm0.001348$ & $[0.652528,0.659223]$ & $0.5967\pm0.0115$ & $0.12351\pm0.00579$ & $0.03954\pm0.00280$ \\
9--19$^{\dagger}$ & $0.658179\pm0.000448$ & $[0.657067,0.659291]$ & $0.5889\pm0.0019$ & $0.11208\pm0.00318$ & $0.03009\pm0.00272$ \\
11--21 & $0.662655\pm0.000623$ & $[0.661106,0.664203]$ & $0.5944\pm0.0077$ & $0.09177\pm0.00340$ & $0.02014\pm0.00160$ \\
14--24 & $0.672529\pm0.000999$ & $[0.670049,0.675010]$ & $0.5878\pm0.0241$ & $0.05508\pm0.00292$ & $0.00756\pm0.00102$ \\
17--27 & $0.682186\pm0.000523$ & $[0.680887,0.683485]$ & $0.5722\pm0.0139$ & $0.02993\pm0.00135$ & $0.00329\pm0.00040$ \\
21--31 & $0.681695\pm0.000585$ & $[0.680240,0.683149]$ & $0.5522\pm0.0107$ & $0.03134\pm0.00111$ & $0.00550\pm0.00040$ \\
\bottomrule
\multicolumn{6}{l}{$^{\star}$ minimum mean; $^{\dagger}$ frozen representative. Basin: 3--13, 6--16, 9--19.}
\end{tabular*}
\end{table*}

\subsection{Matched Capacity and Scale}

\begin{table*}[t!]
\centering\footnotesize
\caption{Controlled three-seed dev metrics for the reference models.}
\label{tab:matched_objective_full}
\begin{tabular*}{\textwidth}{@{\extracolsep{\fill}}llrrrrr@{}}
\toprule
\textbf{Model} & \textbf{Allocation} & \textbf{Trainable M} & \textbf{Dev loss} & \textbf{Pref. acc.} & \textbf{Margin} & \textbf{Chosen KL} \\
\midrule
Qwen2.5-7B & Middle/Selected & 3.244 & $0.676943\pm0.000640$ & $0.5822$ & $0.03911$ & $0.00634$ \\
             & Matched & 3.244 & $0.685571\pm0.000989$ & $0.5611$ & $0.01748$ & $0.00253$ \\
             & All & 10.093 & $0.674154\pm0.000980$ & $0.5944$ & $0.04830$ & $0.00913$ \\
Llama-3.1-8B & Middle & 4.686 & $0.662655\pm0.000623$ & $0.5944$ & $0.09177$ & $0.02014$ \\
              & Matched & 4.686 & $0.668384\pm0.001575$ & $0.5889$ & $0.07464$ & $0.02038$ \\
              & All & 13.631 & $0.654588\pm0.002268$ & $0.5956$ & $0.15084$ & $0.06002$ \\
\bottomrule
\end{tabular*}
\end{table*}

Paired seed differences confirm that Middle improves on Matched for Qwen ($-0.0086$, 95\% CI $[-0.0095,-0.0078]$) and Llama ($-0.0057$, $[-0.0109,-0.0006]$). Selected versus All is unresolved for Qwen ($+0.0028$, $[-0.0010,+0.0066]$) and Llama ($+0.0036$, $[-0.0024,+0.0096]$).

\begin{table*}[t!]
\centering\footnotesize
\caption{Qwen scale results. Objective effects are three-seed dev-loss differences; behavior effects are Qwen3-Max Overall differences from the seed-42 generations. Effects are Middle minus All.}
\label{tab:scale_full}
\begin{tabular*}{0.90\textwidth}{@{\extracolsep{\fill}}lrrr@{}}
\toprule
\textbf{Scale} & \textbf{Middle dev loss} & \textbf{All dev loss} & \textbf{Middle$-$All [95\% CI]} \\
\midrule
3B  & $0.674580\pm0.000772$ & $0.667378\pm0.000748$ & $+0.0072\;[+0.0035,+0.0109]$ \\
14B & $0.663891\pm0.001124$ & $0.662537\pm0.001771$ & $+0.0014\;[-0.0058,+0.0085]$ \\
\midrule
\multicolumn{4}{l}{\emph{Qwen3-Max behavioral effect (Middle$-$All)}} \\
3B  & \multicolumn{2}{l}{$-0.005$} & $[-0.063,+0.053]$ \\
14B & \multicolumn{2}{l}{$+0.001$} & $[-0.049,+0.051]$ \\
\bottomrule
\end{tabular*}
\end{table*}

\section{Blind Human Evaluation}
\label{sec:app_human}

The human study compares the frozen Selected window with All for Qwen2.5-7B and Llama-3.1-8B. For each comparison independently, 120 source IDs are sampled without replacement from the aligned 180-prompt human pool using a deterministic pseudorandom seed constructed from global seed 4242, the comparison code, and \texttt{formal-sample}; sampled IDs are then sorted. The Qwen and Llama formal sets can therefore differ. Each comparison has three independent annotators and six repeated quality-control prompts per annotator. Model identity, allocation, and response orientation are hidden. The public sheets contain randomized LEFT/RIGHT/TIE choices; private keys restore the allocation identity after annotation.

Annotators read Chinese translations of the English prompts and outputs. A temperature-zero \texttt{deepseek-chat} translation instruction requires faithful translation with no summarization, correction, comparison, or source guessing. Identical source strings always receive the same canonical translation. Across all six packs, 2{,}268 text cells map to 623 unique source strings with no conflicting translations. This design evaluates preferences over translated renderings, so stylistic and semantic changes introduced by translation remain a limitation.

All 756 submitted choices are valid. The formal estimate excludes the 36 repeated QC rows and retains them only for self-consistency. Confidence intervals use 100{,}000 prompt-cluster bootstrap samples with seed 42; the three votes from each prompt are resampled together.

\begin{table*}[t!]
\centering\footnotesize
\setlength{\tabcolsep}{2.5pt}
\caption{Complete blind human-evaluation results. A is Selected and B is All. Net preference includes ties and is $(A-B)/360$.}
\label{tab:human_full}
\begin{adjustbox}{max width=\textwidth}
\begin{tabular}{lrrrrrr}
\toprule
\textbf{Model} & \textbf{Raw A/Tie/B} & \textbf{Majority A/Tie/B} & \textbf{Net A$-$B [95\% CI]} & \textbf{A win, no ties [CI]} & \textbf{Fleiss $\kappa$} & \textbf{Repeat consistency} \\
\midrule
Qwen2.5-7B & 114/100/146 & 38/39/43 & $-0.089\;[-0.203,+0.028]$ & $0.438\;[0.358,0.519]$ & 0.401 & 0.833 \\
Llama-3.1-8B & 133/69/158 & 41/29/50 & $-0.069\;[-0.197,+0.061]$ & $0.457\;[0.378,0.537]$ & 0.514 & 0.889 \\
\bottomrule
\end{tabular}
\end{adjustbox}
\end{table*}

Both point estimates favor All, and both intervals include zero. Leave-one-annotator-out estimates preserve the same direction. Three-way agreement is defined as an exact prompt-level match between the human-majority A/B/tie label and the order-swapped LLM judge's aggregated A/B/tie label. It is higher for Llama; evaluator disagreement remains a limitation.

\section{Trained-Adapter Interventions}
\label{sec:app_interventions}

Interventions operate on the seed-42 All checkpoint after training. A removal intervention masks adapter contributions from one depth region while retaining all other trained adapters. A retention intervention keeps contributions from only one region. Teacher-forced effects compare the intervened model with intact All on the same 300 dev prompts; behavioral effects use Qwen3-Max on 299--300 paired generations.

\begin{table*}[t!]
\centering\footnotesize
\caption{Complete intervention effects for removal experiments. Positive dev-loss effects indicate worse objective fit after removal; negative Qwen3-Max effects indicate lower Overall.}
\label{tab:intervention_full}
\begin{tabular*}{\textwidth}{@{\extracolsep{\fill}}lllrl@{}}
\toprule
\textbf{Model} & \textbf{Removed adapters} & \textbf{Metric} & \textbf{Effect [95\% CI]} & \textbf{$n$} \\
\midrule
Qwen2.5-7B & Shallow & dev DPO loss & $-0.0006\;[-0.0057,+0.0043]$ & 300 \\
Qwen2.5-7B & Middle & dev DPO loss & $+0.0124\;[+0.0050,+0.0194]$ & 300 \\
Qwen2.5-7B & Deep & dev DPO loss & $+0.0038\;[+0.0006,+0.0069]$ & 300 \\
Llama-3.1-8B & Shallow & dev DPO loss & $+0.0110\;[-0.0022,+0.0236]$ & 300 \\
Llama-3.1-8B & Middle & dev DPO loss & $+0.0157\;[+0.0032,+0.0278]$ & 300 \\
Llama-3.1-8B & Deep & dev DPO loss & $+0.0021\;[-0.0023,+0.0066]$ & 300 \\
Llama-3.1-8B & Selected 9--19 & dev DPO loss & $+0.0207\;[+0.0044,+0.0364]$ & 300 \\
\midrule
Qwen2.5-7B & Middle & Qwen3-Max Overall & $+0.0000\;[-0.0967,+0.1000]$ & 300 \\
Qwen2.5-7B & Shallow & Qwen3-Max Overall & $-0.1000\;[-0.2067,+0.0067]$ & 300 \\
Qwen2.5-7B & Deep & Qwen3-Max Overall & $+0.0700\;[-0.0230,+0.1670]$ & 300 \\
Llama-3.1-8B & Middle & Qwen3-Max Overall & $-0.0301\;[-0.1538,+0.0903]$ & 299 \\
Llama-3.1-8B & Selected 9--19 & Qwen3-Max Overall & $-0.2575\;[-0.3946,-0.1204]$ & 299 \\
Llama-3.1-8B & Shallow & Qwen3-Max Overall & $-0.4548\;[-0.6054,-0.3110]$ & 299 \\
Llama-3.1-8B & Deep & Qwen3-Max Overall & $+0.0167\;[-0.0736,+0.1070]$ & 299 \\
\bottomrule
\end{tabular*}
\end{table*}

\begin{table*}[t!]
\centering\footnotesize
\caption{Complete retention interventions. Fractions are descriptive ratios of the Base-to-All dev-loss improvement preserved when only the named trained adapters remain.}
\label{tab:intervention_retention_full}
\begin{tabular*}{\textwidth}{@{\extracolsep{\fill}}llrrr@{}}
\toprule
\textbf{Model} & \textbf{Retained adapters} & \textbf{Dev DPO loss} & \textbf{$\Delta$ loss vs. All [95\% CI]} & \textbf{Gain retained} \\
\midrule
Qwen2.5-7B & Shallow & 0.690307 & $+0.015197\;[+0.006988,+0.023018]$ & 0.157 \\
            & Middle  & 0.679340 & $+0.004230\;[-0.001447,+0.009677]$ & 0.765 \\
            & Deep    & 0.687899 & $+0.012790\;[+0.002875,+0.022162]$ & 0.291 \\
\midrule
Llama-3.1-8B & Shallow        & 0.673010 & $+0.020347\;[+0.005990,+0.034291]$ & 0.497 \\
              & Middle         & 0.667030 & $+0.014368\;[-0.000187,+0.028303]$ & 0.645 \\
              & Deep           & 0.687986 & $+0.035323\;[+0.012951,+0.057318]$ & 0.127 \\
              & Selected 9--19 & 0.660704 & $+0.008041\;[-0.003422,+0.018608]$ & 0.801 \\
\bottomrule
\end{tabular*}
\end{table*}

Among the fixed depth thirds, Middle-only retention preserves 0.765 of the Qwen All-over-Base dev-loss gain and 0.645 of the Llama gain, the largest fraction for each model. Llama Selected-only retention preserves 0.801 for the frozen 9--19 window. These results establish dependence on trained adapter components. Simultaneously trained blocks can interact nonlinearly, so the intervention does not identify an exclusive function for any block.

\section{Implementation-Level Efficiency}
\label{sec:app_efficiency}

The profiling run uses the seed-42 training configuration. Adapter size and peak allocated CUDA memory are measured. Adapter tensors are float32; one float32 gradient per trainable parameter and two float32 Adam moments make the reported gradient and optimizer-state quantities exact. Step time and tokens per second use timestamp and token deltas from steps 50 through 600, excluding setup and warm-up.

\begin{table*}[t!]
\centering\scriptsize
\setlength{\tabcolsep}{3pt}
\caption{Complete NVIDIA H800 efficiency accounting. All runs execute a full forward pass through the base model.}
\label{tab:efficiency_full}
\begin{tabular*}{\textwidth}{@{\extracolsep{\fill}}llrrrrrrr@{}}
\toprule
\textbf{Model} & \textbf{Allocation} & \textbf{Trainable M} & \textbf{Adapter MiB} & \textbf{Grad MiB} & \textbf{Adam MiB} & \textbf{Peak GiB} & \textbf{s/step} & \textbf{tok/s} \\
\midrule
Qwen2.5-7B & Middle & 3.244 & 23.29 & 12.38 & 24.75 & 59.73 & 1.315 & 5422.4 \\
 & Matched & 3.244 & 23.31 & 12.38 & 24.75 & 77.30 & 1.615 & 4415.8 \\
 & All & 10.093 & 49.43 & 38.50 & 77.00 & 77.44 & 1.585 & 4498.3 \\
Llama-3.1-8B & Middle & 4.686 & 34.31 & 17.88 & 35.75 & 58.37 & 1.452 & 4904.2 \\
 & Matched & 4.686 & 34.33 & 17.88 & 35.75 & 76.42 & 1.796 & 3965.5 \\
 & All & 13.631 & 68.45 & 52.00 & 104.00 & 76.59 & 1.770 & 4023.2 \\
\bottomrule
\end{tabular*}
\end{table*}

Middle and Matched have equal trainable counts and optimizer-state accounting. Their measured peak-memory and step-time differences arise in this implementation and hardware environment. The model still executes every frozen base layer, so the table does not establish a general training-time speedup.

\section{Exploratory Analyses from an Earlier Pipeline}
\label{sec:app_exploratory}

The analyses in this section use an earlier pipeline or single-seed diagnostics. They provide scope checks and mechanistic context, and none enters the controlled headline comparisons.

\subsection{Layer-Wise Diagnostics}
\label{sec:app_mechanism}

\begin{figure*}[t!]
    \centering
    \includegraphics[width=\linewidth]{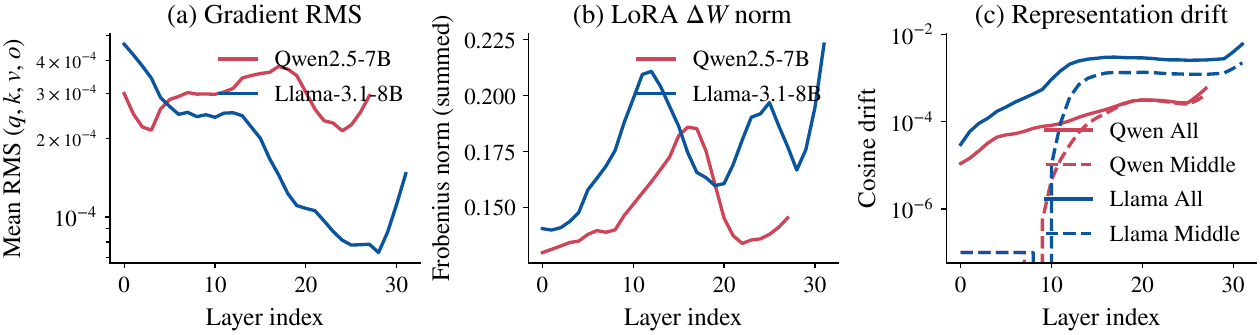}
    \caption{Correlational diagnostics along depth. (a) Per-layer gradient RMS at recorded step 95 of All training. (b) Layer-summed Frobenius norm of LoRA $\Delta W$ after All training. (c) Per-layer cosine drift between base and aligned hidden states on 256 held-out prompts; solid lines show All and dashed lines show Middle. Shading covers the union of the two models' middle thirds.}
    \label{fig:mechanism}
\end{figure*}

Gradient RMS is recorded every five steps from step 0 through step 95. For each layer, the plotted value averages the four $q,k,v,o$ projections in the final record. Update norms use $\Delta W=(\alpha/r)BA$ and sum the four projections in each layer. Representation drift uses the last-token hidden state on 256 held-out dev prompts and reports $1-\cos(h_l^{\mathrm{base}},h_l^{\mathrm{aligned}})$.

\begin{table*}[t!]
\centering\footnotesize
\caption{Qualitative summary of the correlational depth diagnostics.}
\label{tab:mechanism_summary_clean}
\begin{tabularx}{\textwidth}{lXX}
\toprule
\textbf{Signal} & \textbf{Qwen2.5-7B} & \textbf{Llama-3.1-8B} \\
\midrule
Gradient RMS & Peak at layers 14--17; lower values in late layers & Broad early-to-middle plateau; drop after layer 21 \\
LoRA update norm & Middle peak & Middle peak with a secondary late cluster \\
Representation drift & Rises from middle depth and propagates downstream & Jumps near middle entry and propagates downstream \\
\bottomrule
\end{tabularx}
\end{table*}

These diagnostics are correlational. The trained-adapter interventions in \S\ref{sec:app_interventions} provide the direct component-dependence tests.

\subsection{Qwen3-8B Headroom Boundary}
\label{sec:app_qwen3_headroom}

Qwen3-8B \citep{yang2025qwen3} is retained only as a headroom boundary case from the earlier pipeline. All-layer LoRA-DPO lowers teacher-forced DPO loss from the unadapted reference value $\ln 2=0.693147$ to 0.678554, while Qwen3-Max Overall is 3.264 for All and 3.306 for Base. Order-swapped pairwise checks on the same 1{,}000 stored prompts find no reliable All-versus-Base preference: DeepSeek-v4-pro gives $+0.009$ ($[-0.044,+0.061]$) and GPT-5.5 gives $+0.022$ ($[-0.035,+0.078]$). Because all-layer tuning itself has no detectable behavioral gain in this case, Qwen3-8B does not support an inference about the ordering of layer allocations.

\subsection{HelpSteer2 Transfer}

The exploratory HelpSteer2 run uses 6{,}766 non-tie training pairs and 300 held-out validation prompts. Its effective batch is 16 with micro-batch 2, accumulation 8, and gradient checkpointing. Table~\ref{tab:helpsteer_compact} shows that the training-loss ordering transfers more clearly than the Qwen3-Max behavioral ordering.

The behavioral results do not show an absolute gain over Base for either model. On Qwen, Global (3.047) also exceeds Middle (2.880), reversing the corresponding HH-RLHF direction. We therefore treat HelpSteer2 as a boundary case for dataset and rubric transfer, rather than evidence that a fixed behavioral ordering generalizes.

\subsection{TruthfulQA and SimPO Checks}

TruthfulQA uses 817 multiple-choice questions in zero-generation log-likelihood mode. All 95\% intervals for the MC1 change relative to Base include zero.

An exploratory SimPO check holds the Qwen2.5-7B data and LoRA configuration fixed ($\mathrm{lr}=5\times10^{-7}$, $\beta=2.0$, $\gamma=1.0$). Qwen3-Max gives Middle minus All as $-0.020$ with 95\% CI $[-0.067,+0.029]$; DeepSeek yields 22.6\%/25.2\% Middle/All wins and 52.2\% ties. Both comparisons are unresolved.

\begin{table}[t!]
\centering\footnotesize
\setlength{\tabcolsep}{2pt}
\caption{Exploratory HelpSteer2 results. Window denotes the earlier sliding-window choice.}
\label{tab:helpsteer_compact}
\begin{adjustbox}{max width=\linewidth}
\begin{tabular}{lrrrr}
\toprule
& \multicolumn{2}{c}{\textbf{Qwen2.5-7B}} & \multicolumn{2}{c}{\textbf{Llama-3.1-8B}} \\
\cmidrule(lr){2-3}\cmidrule(lr){4-5}
\textbf{Allocation} & \textbf{Train loss} & \textbf{Overall} & \textbf{Train loss} & \textbf{Overall} \\
\midrule
Base & --- & 3.067 & --- & 1.680 \\
All & 0.6621 & 2.903 & 0.6532 & 1.613 \\
Middle & 0.6689 & 2.880 & 0.6663 & 1.582 \\
Window & 0.6671 & 2.933 & 0.6584 & 1.577 \\
Global & 0.6929 & 3.047 & 0.6888 & 1.680 \\
\bottomrule
\end{tabular}
\end{adjustbox}
\end{table}

\begin{table}[t!]
\centering\footnotesize
\setlength{\tabcolsep}{2pt}
\caption{Exploratory Qwen2.5-7B TruthfulQA results.}
\label{tab:truthfulqa_clean}
\begin{adjustbox}{max width=\linewidth}
\begin{tabular}{lrrr}
\toprule
\textbf{Allocation} & \textbf{MC1} & \textbf{MC2} & \textbf{$\Delta$ MC1 vs. Base [95\% CI]} \\
\midrule
Base & 0.3097 & 0.4207 & --- \\
Local & 0.3146 & 0.4205 & $+0.0049\;[-0.0024,+0.0122]$ \\
Middle & 0.3158 & 0.4203 & $+0.0061\;[-0.0012,+0.0135]$ \\
Global & 0.3133 & 0.4203 & $+0.0037\;[-0.0012,+0.0098]$ \\
All & 0.3146 & 0.4198 & $+0.0049\;[-0.0012,+0.0122]$ \\
\bottomrule
\end{tabular}
\end{adjustbox}
\end{table}

\FloatBarrier